\documentclass{article}
\usepackage{a4wide}
\def\sref#1{\S\ref{#1}}
\def\code#1{{\tt #1}}
\title{Transducers from Rewrite Rules with Backreferences}
\author{\begin{tabular}{cc}
Dale Gerdemann & Gertjan van Noord\\
University of Tuebingen & Groningen University\\
Kl. Wilhelmstr. 113 & PO Box 716\\
D-72074 Tuebingen & NL 9700 AS Groningen\\
{\tt dg@sfs.nphil.uni-tuebingen.de} & {\tt vannoord@let.rug.nl}
\end{tabular}
}
\begin{document}
\maketitle
\begin{abstract}
  
  Context sensitive rewrite rules have been widely used in several
  areas of natural language processing, including syntax, morphology,
  phonology and speech processing. Kaplan and Kay, Karttunen, and
  Mohri \& Sproat have given various algorithms to compile such
  rewrite rules into finite-state transducers. The present paper
  extends this work by allowing a limited form of backreferencing in
  such rules. The explicit use of backreferencing leads to more
  elegant and general solutions.
\end{abstract}

\section{\label{intro}Introduction}

Context sensitive rewrite rules have been widely used in several areas
of natural language processing. Johnson \cite{john:form72} has
shown that such rewrite rules are equivalent to finite state
transducers in the special case
that they are not allowed to rewrite their own output. An algorithm
for compilation into transducers was provided by
\cite{kapl:regu94}. Improvements and extensions to this algorithm
have been provided by \cite{kart:95}, \cite{kart:repl97},
\cite{karttunen:96} and \cite{mohri-sproat:96}. In this paper,
the algorithm will be extended to provide a limited form of
backreferencing.  Backreferencing has been implicit in previous
research, such as in the ``batch rules'' of \cite{kapl:regu94},
bracketing transducers for finite-state parsing \cite{karttunen:96},
and the ``LocalExtension'' operation of
\cite{roche_schabes95}. The explicit use of
backreferencing leads to more elegant and general solutions.

Backreferencing is widely used in editors, scripting languages and
other tools employing regular expressions \cite{frie:mast97}.  For
example, Emacs uses the special brackets \verb+\(+ and \verb+\)+ to
capture strings along with the notation \verb+\+$n$ to recall the
$n$th such string. The expression {\bf \verb+\+(a*\verb+\+)b\verb+\+1}
matches strings of the form $a^nba^n$. Unrestricted use of
backreferencing thus can introduce non-regular languages. For NLP
finite state calculi \cite{kart:regu96,noord:fsa97} this is
unacceptable.  The form of backreferences introduced in this paper
will therefore be restricted.

The central case of an allowable backreference is:

\begin{equation} \label{base}
x\Rightarrow  T(x)/\lambda\mbox{\_\_}\rho
\end{equation}

\noindent This says that each string $x$ preceded by $\lambda$ and
followed by $\rho$ is replaced by $T(x)$, where $\lambda$ and $\rho$
are arbitrary regular expressions, and $T$ is a
transducer.\footnote{The syntax at this point is merely
  suggestive. As an example, suppose that $T_{acr}$ transduces
  phrases into acronyms. Then 
\[x\Rightarrow T_{acr}(x)/\langle
\mbox{abbr}\rangle\mbox{\_\_}\langle /\mbox{abbr}\rangle\] 
would transduce {\tt <abbr>non-deterministic finite automaton</abbr>}
into {\tt <abbr>NDFA</abbr>}.

To compare this with a backreference in Perl, suppose that
$T_{acr}$ is a subroutine that converts phrases into acronyms and
that $R_{acr}$ is a regular expression matching phrases that can
be converted into acronyms. Then (ignoring the left context) one can
write something like:
s/($R_{acr}$)(?=$\langle$/ABBR$\rangle$)/$T_{acr}$(\$1)/ge;.
The backreference variable, \$1, will be set to whatever string
$R_{acr}$ matches.} This contrasts sharply with the rewriting
rules that follow the tradition of Kaplan \& Kay:

\begin{equation} \label{noback}
\phi\Rightarrow  \psi/\lambda\mbox{\_\_}\rho
\end{equation}

\noindent In this case, any string from the language $\phi$ is replaced
by any string independently chosen from the language $\psi$.

We also allow multiple (non-permuting)
backreferences of the form:

\begin{equation}\label{multi}
x_1 x_2\ldots x_n \Rightarrow  T_1(x_1) T_2(x_2) \ldots
T_n(x_n)/\lambda\mbox{\_\_}\rho
\end{equation}

\noindent Since transducers are closed under concatenation, 
handling multiple backreferences  reduces to the problem
of handling a single backreference:

\begin{equation}
x\Rightarrow  (T_1 \cdot T_2 \cdot \ldots \cdot T_n)(x)/\lambda\mbox{\_\_}\rho
\end{equation}

\noindent A problem arises if we want capturing to follow the
POSIX standard requiring a longest-capture strategy.  Friedl
\cite{frie:mast97} (p.\/ 117), for example, discusses matching the
regular expression (to$\mid$top)(o$\mid$polo)?(gical$\mid$o?logical)
against the word: {\bf topological}. The desired result is that (once
an overall match is established) the first set of parentheses should
capture the longest string possible ({\bf top}); the second set should
then match the longest string possible from what's left ({\bf o}), and
so on.  Such a left-most longest match concatenation operation is
described in \sref{topo}.

In the following section, we initially concentrate on the simple case
in (\ref{base}) and show how (\ref{base}) may be compiled assuming
left-to-right processing along with the overall longest match strategy
described by \cite{karttunen:96}.

The major components of the algorithm are not new, but straightforward
modifications of components presented in \cite{karttunen:96} and
\cite{mohri-sproat:96}.  We improve upon existing approaches
because we solve a problem concerning the use of special marker
symbols (\sref{alpha}).  A further contribution is that all steps are
implemented in a freely available system, the FSA Utilities of
\cite{noord:fsa97} (\sref{fsu}).

\section{The Algorithm \label{algorithm}}

\subsection{Preliminary Considerations}

Before presenting the algorithm proper, we will deal with a couple of
meta issues. First, we introduce our version of the finite state
calculus in \sref{fsu}. The treatment of special marker symbols
is discussed in \sref{alpha}. Then in \sref{util}, we
discuss various utilities that will be essential for the algorithm.

\subsubsection{\label{fsu}FSA Utilities}

\begin{table}
\begin{tabular}{cl}
\tt []                  & empty string \\
\tt [E1,\dots En]    & concatenation of \tt E1 \dots En \\
\tt \verb+{}+           & empty language \\
\tt \verb+{+E1,\dots En\verb+}+    & union of \tt E1,\dots En\\
\tt E*                  & Kleene closure\\
\tt E\verb+^+                  & optionality\\
\tt \~{}E                  & complement\\
\tt E1-E2               & difference\\
\tt \verb+$+ E          & containment\\ 
\tt E1 \verb+&+ E2      & intersection\\
\tt ?                   & any symbol\\
\tt A:B                 & pair\\
\tt E1 x E2             & cross-product\\
\tt A o B               & composition\\
\tt domain(E)           & domain of a transduction\\
\tt range(E)            & range of a transduction\\
\tt identity(E)         & identity transduction\\
\tt inverse(E)          & inverse transduction\\
\end{tabular}
\centering
\caption{\label{notation}Regular expression operators.}
\end{table}

The algorithm is implemented in the FSA Utilities \cite{noord:fsa97}.
We use the notation provided by the toolbox throughout this paper.
Table~\ref{notation} lists the relevant regular expression operators.
FSA Utilities offers the possibility to define new regular expression
operators.  For example, consider the definition of the nullary
operator {\tt vowel} as the union of the five vowels:
\begin{verbatim}
macro(vowel,{a,e,i,o,u}).
\end{verbatim}
In such macro definitions, Prolog variables can be used in order to
define new n-ary regular expression operators in terms of existing
operators. For instance, the {\em lenient\_composition} operator 
\cite{karttunen:98} is defined by:
\begin{verbatim}
macro(priority_union(Q,R), 
      {Q, ~domain(Q) o R}).
macro(lenient_composition(R,C), 
      priority_union(R o C,R)).
\end{verbatim}
Here, \code{priority\_union} of two regular
expressions \code{Q} and \code{R} is defined as the union of \code{Q}
and the composition of the complement of the domain of \code{Q} with
\code{R}. Lenient composition of  \code{R} and \code{C}
is defined as the priority union of the composition of \code{R} and
\code{C} (on the one hand) and \code{R} (on the other hand).

Some operators, however, require something more than simple macro
expansion for their definition. For example, suppose a user wanted
to match $n$ occurrences of some pattern. The FSA Utilities already
has the '*' and '+' quantifiers, but any other operators like this
need to be user defined. For this purpose, the FSA Utilities
supplies simple Prolog hooks allowing this general quantifier to be
defined as:

\begin{verbatim}
macro(match_n(N,X),Regex) :-
   match_n(N,X,Regex).

match_n(0,_X,[]).
match_n(N,X,[X|Rest]) :-
   N > 0,
   N1 is N-1,
   match_n(N1,X,Rest).
\end{verbatim}

\noindent For example: \verb+match_n(3,a)+ is equivalent to the
ordinary finite state calculus expression \verb+[a,a,a]+.

Finally, regular expression operators can be defined in terms of
operations on the underlying automaton. In such cases, Prolog hooks
for manipulating states and transitions may be used.
This functionality has been used in \cite{noord-gerd99} to provide
an implementation of the algorithm in \cite{mohri-sproat:96}.

\subsubsection{\label{alpha}Treatment of Markers}

Previous algorithms for compiling rewrite rules into transducers have
followed  \cite{kapl:regu94} by
introducing special marker symbols ({\em markers}) 
into strings in order to mark off
candidate regions for replacement. The assumption is that these
markers are outside the resulting transducer's alphabets. But
previous algorithms have not ensured that the assumption holds.

This problem was recognized by \cite{karttunen:96}, whose algorithm
starts with a filter transducer which filters out any string
containing a marker.  This is problematic for two reasons. First, when
applied to a string that does happen to contain a marker, the
algorithm will simply fail. Second, it leads to logical problems in
the interpretation of complementation. Since the complement of a
regular expression R is defined as $\Sigma-R$, one needs to know
whether the marker symbols are in $\Sigma$ or not. This has not been
clearly addressed in previous literature.

We have taken a different approach by providing a contextual way of
distinguishing markers from non-markers.  Every symbol used in the
algorithm is replaced by a pair of symbols, where the second member of
the pair is either a 0 or a 1 depending on whether the first member is
a marker or not.\footnote{This approach is
  similar to the idea of laying down tracks as in the compilation of
  monadic second-order logic into automata
  \nocite{klar:mona97}Klarlund (1997, p. 5). In
  fact, this technique could possibly be used for a more efficient
  implementation of our algorithm: instead of adding transitions over
  0 and 1, one could represent the alphabet as bit sequences and then
  add a final 0 bit for any ordinary symbol and a final 1 bit for a
  marker symbol. } 
As the first step in the algorithm, 0's are inserted
after every symbol in the input string to indicate that initially
every symbol is a non-marker. This is defined as:
\begin{verbatim}
macro(non_markers,[?,[]:0]*).
\end{verbatim}

Similarly, the following macro can be used to insert a 0 after every
symbol in an arbitrary expression E.

\begin{verbatim}
macro(non_markers(E),
      range(E o non_markers)).
\end{verbatim}
\noindent
Since \code{E} is a recognizer, it is first coerced to
\code{identity(E)}. This form of implicit conversion is standard in
the finite state calculus.

Note that 0 and 1 are perfectly ordinary alphabet
symbols, which may also be used within a replacement. For example, the
sequence [1,0] represents a non-marker use of the symbol 1.

\subsubsection{\label{util}Utilities}

Before describing the algorithm, it will be helpful to have at our
disposal a few general tools, most of which were described already in
\cite{kapl:regu94}.  These tools, however, have been modified so
that they work with our approach of distinguishing markers from
ordinary symbols.  So to begin with, we provide macros to describe the
alphabet and the alphabet extended with marker symbols:

\begin{verbatim}
macro(sig,[?,0]).
macro(xsig,[?,{0,1}]).
\end{verbatim}

The macro \code{xsig} is useful for defining a specialized version of
complementation and containment:

\begin{verbatim}
macro(not(X),xsig* - X).
macro($$(X),[xsig*,X,xsig*]).
\end{verbatim}

The algorithm uses four kinds of brackets, so it will be convenient to
define macros for each of these brackets, and for a few disjunctions.

\begin{verbatim}
macro(lb1,['<1',1]).
macro(lb2,['<2',1]).
macro(rb2,['2>',1]).
macro(rb1,['1>',1]).
macro(lb,{lb1,lb2}).
macro(rb,{rb1,rb2}).
macro(b1,{lb1,rb1}).
macro(b2,{lb2,rb2}).
macro(brack,{lb,rb}).
\end{verbatim}

As in Kaplan \& Kay, we define an Intro(S) operator that
produces a transducer that freely introduces instances of 
S into an input string. We extend this idea to create
a family of Intro operators. It is often the case that we want to
freely introduce marker symbols into a string at any position {\em
  except} the beginning or the end.

\begin{verbatim}
%% Free introduction
macro(intro(S),{xsig-S,[] x S}*).

%% Introduction, except at begin
macro(xintro(S),{[],[xsig-S,intro(S)]}).

%% Introduction, except at end
macro(introx(S),{[],[intro(S),xsig-S]}).

%% Introduction, except at begin & end
macro(xintrox(S),{[],[xsig-S],
   [xsig-S,intro(S),xsig-S]}).
\end{verbatim}

This family of Intro operators is useful for defining a family
of Ignore operators:

\begin{verbatim}
macro( ign( E1,S),range(E1 o  intro( S))).
macro(xign( E1,S),range(E1 o xintro( S))).
macro( ignx(E1,S),range(E1 o  introx(S))).
macro(xignx(E1,S),range(E1 o xintrox(S))).
\end{verbatim}

In order to create filter transducers to ensure that markers are
placed in the correct positions, Kaplan \& Kay introduce the operator
\code{P-iff-S(L1,L2)}. A string is described by this expression iff
each prefix in \code{L1} is followed by a suffix in \code{L2} and each
suffix in \code{L2} is preceded by a prefix in \code{L1}. In our
approach, this is defined as:

\begin{verbatim}
macro(if_p_then_s(L1,L2), 
      not([L1,not(L2)])).
macro(if_s_then_p(L1,L2), 
      not([not(L1),L2])).
macro(p_iff_s(L1,L2), 
      if_p_then_s(L1,L2) 
            & 
      if_s_then_p(L1,L2)).
\end{verbatim}

To make the use of \code{p\_iff\_s} more convenient, we introduce a new
operator \code{l\_iff\_r(L,R)}, which describes 
strings where every string position is preceded by a string in
\code{L} just in case it is followed by a string in \code{R}:

\begin{verbatim}
macro(l_iff_r(L,R), 
   p_iff_s([xsig*,L],[R,xsig*])).
\end{verbatim}

Finally, we introduce a new operator \code{if(Condition,Then,Else)} for
conditionals. This operator is extremely useful, but in order for it
to work within the finite state calculus, one needs a convention as to
what counts as a boolean true or false for the condition
argument.  It is possible to define {\tt true} as the universal
language and {\tt false} as the empty language:

\begin{verbatim}
macro(true,? *).   macro(false,{}).
\end{verbatim}

With these definitions, we can use the complement operator as
negation, the intersection operator as conjunction and the union
operator as disjunction. Arbitrary expressions may be coerced to
booleans using the following macro:

\begin{verbatim}
macro(coerce_to_boolean(E), 
      range(E o (true x true))).
\end{verbatim}

\noindent Here, \code{E} should describe a recognizer. 
\code{E} is composed with the universal transducer, which transduces from
anything (\code{?*}) to anything (\code{?*}). 
Now with this background, we can define
the conditional:

\begin{verbatim}
macro(if(Cond,Then,Else), 
   {  coerce_to_boolean(Cond) o Then,
     ~coerce_to_boolean(Cond) o Else
   }).
\end{verbatim}

\subsection{Implementation}

A rule of the form  $x \rightarrow T(x)/\lambda\mbox{\_\_}\rho$
will be written as \code{replace(T,Lambda,Rho)}. Rules of the more general
form $x_1 \ldots x_n \Rightarrow  T_1(x_1) \ldots
T_n(x_n)/\lambda\mbox{\_\_}\rho$ will be discussed in \sref{topo}. The
algorithm consists of nine steps composed as in figure~\ref{replace}.

\begin{figure*}
\begin{verbatim}
macro(replace(T,Left,Right),
            non_markers             % introduce 0 after every symbol
                o                   % (a b c => a 0 b 0 c 0).
             r(Right)               % introduce rb2 before any string
                o                   % in Right.
            f(domain(T))            % introduce lb2 before any string in
                o                   % domain(T) followed by rb2.
       left_to_right(domain(T))     % lb2 ... rb2 around domain(T) optionally
                o                   % replaced by lb1 ... rb1
       longest_match(domain(T))     % filter out non-longest matches marked
                o                   % in previous step.
          aux_replace(T)            % perform T's transduction on regions marked
                o                   % off by b1's.
             l1(Left)               % ensure that lb1 must be preceded
                o                   % by a string in Left.
             l2(Left)               % ensure that lb2 must not occur preceded
                o                   % by a string in Left.
         inverse(non_markers)).     % remove the auxiliary 0's.
\end{verbatim}

\caption{\label{replace}Definition of \code{replace} operator.}
\end{figure*}

The names of these steps are mostly derived from \cite{kart:95} and
\cite{mohri-sproat:96} even though the transductions involved are
not exactly the same. In particular, the steps derived from Mohri \&
Sproat (\code{r}, \code{f}, \code{l1} and \code{l2}) will all be
defined in terms of the finite state calculus as opposed to Mohri \&
Sproat's approach of using low-level manipulation of states and
transitions.\footnote{The alternative implementation is provided in \cite{noord-gerd99}.}

The first step, \code{non\_markers}, was already defined above.
For the second step, we first consider a simple special case. If the
empty string is in the language described by \code{Right},
then \code{r(Right)} should insert an
\code{rb2} in every string position. The definition of
\code{r(Right)} is both simpler and more efficient if this is treated
as a special case.  To insert a bracket in every possible string
position, we use:

\begin{verbatim}
[[[] x rb2,sig]*,[] x rb2]
\end{verbatim}

If the empty string is not in \code{Right}, then we must use
\code{intro(rb2)} to introduce the marker \code{rb2}, followed
by \code{l\_iff\_r} to ensure that such markers are immediately
followed by a string in \code{Right}, or more precisely a string
in \code{Right} where additional instances of \code{rb2} are freely
inserted in any position other than the beginning. This expression is
written as:

\begin{verbatim}
            intro(rb2) 
                o 
l_iff_r(rb2,xign(non_markers(Right),rb2))
\end{verbatim}

Putting these two pieces together with the conditional yields:

\begin{verbatim}
macro(r(R),
  if([] & R,       % If: [] is in R:
     [[[] x rb2,sig]*,[] x rb2], 
      intro(rb2)   % Else:
            o
l_iff_r(rb2,xign(non_markers(R),rb2)))).
\end{verbatim}

The third step, \code{f(domain(T))} is implemented as:

\begin{verbatim}
macro(f(Phi), intro(lb2)
                 o
l_iff_r(lb2,[xignx(non_markers(Phi),b2),
             lb2^,rb2])).
\end{verbatim}

The \code{lb2} is first introduced and then, using
\code{l\_iff\_r}, it is constrained to occur immediately before every
instance of (ignoring complexities) \code{Phi} followed by an
\code{rb2}. \code{Phi} needs to be marked as normal text
using \code{non\_markers} and then \code{xign\_x} is used to allow
freely inserted \code{lb2} and \code{rb2} anywhere except at
the beginning and end. The following \code{lb2}\verb+^+ allows an
optional \code{lb2}, which occurs when the empty string is in
\code{Phi}.

The fourth step is a guessing component which (ignoring complexities)
looks for sequences of the form \code{lb2 Phi rb2} and
converts some of these into \code{lb1 Phi rb1}, where the
\code{b1} marking indicates that the sequence is a candidate for
replacement. The complication is that \code{Phi}, as always, must be
converted to \code{non\_markers(Phi)} and instances of \code{b2}
need to be ignored. Furthermore, between pairs of \code{lb1}
and \code{rb1}, instances of \code{lb2} are deleted. These
\code{lb2} markers have done their job and are no longer needed.
Putting this all together, the definition is:

\begin{verbatim}
macro(left_to_right(Phi),
  [[xsig*,
    [lb2 x lb1,
     (ign(non_markers(Phi),b2)
              o
      inverse(intro(lb2))
     ),
     rb2 x rb1]
   ]*, xsig*]).
\end{verbatim}

The fifth step filters out non-longest matches produced in the
previous step. For example (and simplifying a bit), if \code{Phi} is
\code{ab*}, then a string of the form $\ldots$ rb1 a b lb1 b
$\ldots$ should be ruled out since there is an instance of \code{Phi}
(ignoring brackets except at the end) where there is an internal
\code{lb1}. This is implemented as:\footnote{The line with
  \code{\$\$(rb1)} can be optimized a bit: Since we know that an
  \code{rb1} must be preceded by \code{Phi}, we can write:
  \code{[ign\_(non\_markers(Phi),brack),rb1,xsig*])}. This
    may lead to a more constrained (hence smaller) transducer. }

\begin{verbatim}
macro(longest_match(Phi),
  not($$([lb1,
          (ignx(non_markers(Phi),brack)
                     &
                  $$(rb1)
          ),     % longer match must be
          rb     % followed by an rb 
         ]))     % so context is ok
          o
% done with rb2, throw away:
   inverse(intro(rb2))).  
\end{verbatim}

The sixth step performs the transduction described by \code{T}. This
step is straightforwardly implemented, where the main difficulty is
getting \code{T} to apply to our specially marked string:

\begin{verbatim}
macro(aux_replace(T),
  {{sig,lb2},
   [lb1,
    inverse(non_markers) 
         o T o 
       non_markers,
    rb1 x []
   ]
  }*).
\end{verbatim}

The seventh step ensures that \code{lb1} is preceded by a string
in \code{Left}:

\begin{verbatim}
macro(l1(L),
  ign(if_s_then_p(
       ignx([xsig*,non_markers(L)],lb1),
       [lb1,xsig*]),
      lb2)
        o
 inverse(intro(lb1))). 
\end{verbatim}

The eighth step ensures that \code{lb2} is not preceded by a
string in \code{Left}. This is implemented similarly to the
previous step:

\begin{verbatim}
macro(l2(L),
  if_s_then_p(
    ignx(not([xsig*,non_markers(L)]),lb2),
    [lb2,xsig*])
          o
 inverse(intro(lb2))).
\end{verbatim}

Finally the ninth step, \code{inverse(non\_markers)}, removes the
\code{0}'s so that the final result in not marked up in any special
way.

\section{\label{topo}Longest Match Capturing}

As discussed in \sref{intro} the POSIX standard requires that multiple
captures follow a longest match strategy. For multiple captures as in
(\ref{multi}), one establishes first a longest match for
$domain(T_1)\cdot \ldots\cdot domain(T_n)$. Then we
ensure that each of $domain(T_i)$ in turn is
required to match as long as possible, with each one having priority
over its rightward neighbors. To implement this, we define a macro
\code{lm\_concat(Ts)} and use it as:

\begin{verbatim}
replace(lm_concat(Ts),Left,Right)
\end{verbatim}

\begin{figure*}
\small
\begin{verbatim}
macro(lm_concat(Ts),mark_boundaries(Domains) o ConcatTs):-
   domains(Ts,Domains), concatT(Ts,ConcatTs).

domains([],[]).
domains([F|R0],[domain(F)|R]):- domains(R0,R).

concatT([],[]).
concatT([T|Ts], [inverse(non_markers) o T,lb1 x []|Rest]):- concatT(Ts,Rest).

%% macro(mark_boundaries(L),Exp): This is the central component of lm_concat. For our
%% "toplological" example we will have:
%% mark_boundaries([domain([{[t,o],[t,o,p]},[]: #]),
%%                  domain([{o,[p,o,l,o]},[]: #]),
%%                  domain({[g,i,c,a,l],[o^,l,o,g,i,c,a,l]})])
%% which simplifies to:
%% mark_boundaries([{[t,o],[t,o,p]}, {o,[p,o,l,o]}, {[g,i,c,a,l],[o^,l,o,g,i,c,a,l]}]).
%% Then by macro expansion, we get:
%% [{[t,o],[t,o,p]} o non_markers,[]x lb1,
%%  {o,[p,o,l,o]} o non_markers,[]x lb1,
%%  {[g,i,c,a,l],[o^,l,o,g,i,c,a,l]} o non_markers,[]x lb1]
%%              o
%% % Filter 1: {[t,o],[t,o,p]} gets longest match
%% ~ [ignx_1(non_markers({[t,o],[t,o,p]}),lb1),
%%    ign(non_markers({o,[p,o,l,o]}),lb1),
%%    ign(non_markers({[g,i,c,a,l],[o^,l,o,g,i,c,a,l]}),lb1)]
%%                   o
%% % Filter 2: {o,[p,o,l,o]} gets longest match
%% ~ [non_markers({[t,o],[t,o,p]}),lb1,
%%    ignx_1(non_markers({o,[p,o,l,o]}),lb1),
%%    ign(non_markers({[g,i,c,a,l],[o^,l,o,g,i,c,a,l]}),lb1)] 

macro(mark_boundaries(L),Exp):-
   boundaries(L,Exp0), % guess boundary positions 
   greed(L,Exp0,Exp).  % filter non-longest matches

boundaries([],[]).
boundaries([F|R0],[F o non_markers, [] x lb1 |R]):- boundaries(R0,R).

greed(L,Composed0,Composed) :-
   aux_greed(L,[],Filters), compose_list(Filters,Composed0,Composed).

aux_greed([H|T],Front,Filters):- aux_greed(T,H,Front,Filters,_CurrentFilter).

aux_greed([],F,_,[],[ign(non_markers(F),lb1)]).
aux_greed([H|R0],F,Front,[~L1|R],[ign(non_markers(F),lb1)|R1]) :-
   append(Front,[ignx_1(non_markers(F),lb1)|R1],L1),
   append(Front,[non_markers(F),lb1],NewFront),
   aux_greed(R0,H,NewFront,R,R1).

%% ignore at least one instance of E2 except at end
macro(ignx_1(E1,E2), range(E1 o [[? *,[] x E2]+,? +])).

compose_list([],SoFar,SoFar).
compose_list([F|R],SoFar,Composed):- compose_list(R,(SoFar o F),Composed).
\end{verbatim}
\caption{\label{lmconcat}Definition of \code{lm\_concat} operator.}
\end{figure*}

Ensuring the longest overall match is delegated to the \code{replace}
macro, so \code{lm\_concat(Ts)} needs only ensure that each individual
transducer within \code{Ts} gets its proper left-to-right longest
matching priority. This problem is mostly solved by the same
techniques used to ensure the longest match within the \code{replace}
macro. The only complication here is that \code{Ts} can be of
unbounded length. So it is not possible to have a single expression in
the finite state calculus that applies to all possible lenghts. This
means that we need something a little more powerful than mere macro
expansion to construct the proper finite state calculus expression.
The FSA Utilities provides a Prolog hook for this purpose. The
resulting definition of \code{lm\_concat} is given in
figure~\ref{lmconcat}.

Suppose (as in \cite{frie:mast97}), we want to match the following
list of recognizers against the string \code{topological} and insert a
marker in each boundary position. This reduces to applying:

\begin{verbatim}
lm_concat([
   [{[t,o],[t,o,p]},[]: '#'],
   [{o,[p,o,l,o]},[]: '#'],
   {[g,i,c,a,l],[o^,l,o,g,i,c,a,l]}
          ])
\end{verbatim}
  
This expression transduces the string {\tt topological} only to the
string {\tt top\#o\#logical}.\footnote{An anonymous reviewer suggested that
  \code{lm\_concat} could be implemented in the framework of
  \cite{karttunen:96} as: 
\[ \mbox{[t o $\mid$t o p $\mid$ o $\mid$ p o l o ] $\rightarrow$ ... \#}; \]
Indeed the resulting transducer from this
  expression would transduce {\tt topological} into {\tt top\#o\#logical}.
  But unfortunately this transducer would also transduce
  {\tt polotopogical} into {\tt polo\#top\#o\#gical}, since the notion of
  left-right ordering is lost in this expression.}

\section{\label{conclude} Conclusions}

The algorithm presented here has extended previous algorithms for
rewrite rules by adding a limited version of backreferencing. This
allows the output of rewriting to be dependent on the form of the
strings which are rewritten. This new feature brings techniques
used in Perl-like languages into the finite state calculus.  Such an
integration is needed in practical applications where simple
text processing needs to be combined with more
sophisticated computational linguistics techniques.

One particularly interesting example where backreferences are
essential is cascaded deterministic (longest match) finite state
parsing as described for example in Abney \cite{abne:part96} and
various papers in \cite{roche_schabes97}. Clearly, the standard
rewrite rules do not apply in this domain. If {\it NP} is an NP recognizer,
it would not do to say {\it NP} $\Rightarrow [${\it NP}$]/\lambda \mbox{\_\_}\rho$.  Nothing would force the string matched by the
{\it NP} to the left of the arrow to be the same as the string
matched by the {\it NP} to the right of the arrow.

One advantage of using our algorithm for finite state parsing is that
the left and right contexts may be used to bring in top-down
filtering.\footnote{The bracketing operator of \cite{karttunen:96},
  on the other hand, does not provide for left and right contexts.} An
often cited advantage of finite state parsing is robustness. A
constituent is found bottom up in an early level in the cascade even
if that constituent does not ultimately contribute to an S in a later
level of the cascade. While this is undoubtedly an advantage for
certain applications, our approach would allow the introduction of
some top-down filtering while maintaining the robustness of a
bottom-up approach. A second advantage for robust finite state parsing
is that bracketing could also include the notion of ``repair'' as in
\cite{abney:90}. One might, for example, want to say something
like: $\mbox{\it xy} \Rightarrow [_{NP}\, \mbox{\it RepairDet(x)
  RepairN(y)}\,]/\lambda \mbox{\_\_}\rho$\footnote{The syntax here has
  been simplified. The rule should be understood as:
  replace(lm\_concat([[]:'[np', repair\_det, repair\_n,
  []:']'],lambda, rho).} so that an {\it NP} could be parsed as a
slightly malformed {\it Det} followed by a slightly malformed {\it
  N}. RepairDet and RepairN, in this example, could be doing a variety
of things such as: contextualized spelling correction, reordering of
function words, replacement of phrases by acronyms, or any other
operation implemented as a transducer.

Finally, we should mention the problem of complexity. A critical
reader might see the nine steps in our algorithm and conclude that the
algorithm is overly complex. This would be a false conclusion. To
begin with, the problem itself is complex. It is easy to create
examples where the resulting transducer created by any algorithm would
become unmanageably large. But there exist strategies for keeping the
transducers smaller. For example, it is not necessary for all nine
steps to be composed. They can also be cascaded. In that case it will
be possible to implement different steps by different
strategies, e.g.\/ by deterministic or non-deterministic transducers or
bimachines \cite{roch:lang97}.  The range of possibilities
leaves plenty of room for future research.

\bibliographystyle{plain}

\end{document}